# Classification of entities via their descriptive sentences

**Chao Zhao**
Harbin Institute of Technology
zhaochaocs@gmail.com

**Min Zhao**
Baidu Inc.
zhaomin@baidu.com

**Yi Guan**
Harbin Institute of Technology
guanyi@hit.edu.cn

**Abstract**

Hypernym identification of open-domain entities is crucial for taxonomy construction as well as many higher-level applications. Current methods suffer from either low precision or low recall. To decrease the difficulty of this problem, we adopt a classification-based method. We pre-define a concept taxonomy and classify an entity to one of its leaf concept, based on the name and description information of the entity. A convolutional neural network classifier and a K-means clustering module are adopted for classification. We applied this system to 2.1 million Baidu Baike entities, and 1.1 million of them were successfully identified with a precision of 99.36%.

## 1 Introduction

Entity classification aims to determine the type (e.g., person, location) of a certain entity. It is important for several natural language processing (NLP) tasks, such as question answering, textual entailment, machine reading, and text summarization. It is also a crucial resource to introduce the concept-level features to enhance the generalization power of machine learning systems (Paulheim and Fümkranz, 2012).

There are two main methods for entity classification: named entity recognition and classification (NERC) and entity hypernym extraction. The former assumes that the type of an entity can be determined from its context. The classification tags utilized in NER systems are often coarse and do not involve all types of entities. Fine-grained NER faced with the difficulty of creating sufficient training data. The latter method can obtain extremely fine-grained categories of one entity, but it has to rely on an ontology to further obtain more general concepts. It also requires the co-occurrence of the entity and its hypernym in one sentence, which can make the recall lower. If there is no hypernym word in the context of the entity, and the context does not contain much information to determine the classification, these two methods could not work well, and we have to rely on external knowledge sources.

In this paper, we propose another method for entity classification. Rather than inferring the type of an entity from the context, we classify it according to its descriptive sentences. A descriptive sentence is used to describe certain attributes of an entity, and is a natural source for humans to recognize the type of the unknown entity. For example, from the sentence “The logo of Baidu is a bear paw.”, we have no idea what *Baidu* is through the context information. But from a description of *Baidu*:

> Baidu, incorporated on 18 January 2000, is a Chinese web services company

We can easily conclude that *Baidu* is a COMPANY.

Motivated by this idea, we propose a classification-based method for entity classification. We first pre-define a hierarchical concept taxonomy, and then try to classify an entity to a leaf of this taxonomy, based on its description and name information. Our contributions are three-fold:

- We propose a simple classification-based method for entity classification, based on the description and name of the entity.
- We introduce a clustering module to alleviate the data noise and imbalance problem during training, as well as to select the highly confident predicted entities from the prediction set to improve the precision.
- We utilize this architecture on 2.1 million open-domain entities to verify its efficiency. Approximately 1.1 million entities are successfully identified, with a precision of 99.36%.

## 2 Methods

This section presents the dataset and the method used for entity classification.

### 2.1 Dataset and task

The entities and their corresponding descriptions utilized are from Baidu Baike. It contains nearly 15 million Chinese pages until now, which is much larger than that of Wikipedia (nearly 1 million). 12.4 million entities out of 15 have been linked to the Baidu Concept Base, a concept taxonomy, by the features from keywords, tags, or key-value tables, with a precision of 98%. We call them as *known entities*. Another 2.6 million *unknown entities* had not been linked because of the lack of or low quality of these features. The original model is therefore helpless for these entities, yet we want to identify their hypernyms according to their descriptions.

Linking entities to existing concept base can be naturally regarded as a classification task. We first manually select a mini concept taxonomy from the Baidu concept base, and regard the leaf nodes as classification labels. This operation decreases the number of classes to make the classification easier. As shown in left part of Table 1, the mini-taxonomy contains 48 classes. These concepts are selected because they are necessary and fine-grained enough to support subsequent applications. They cover more than 98.6% of the known entities in Baidu Baike.

## 2.2 Pipeline

The workflow of our method is shown in Figure 1, which contains three main modules: pre-processing, model training, and post-processing.

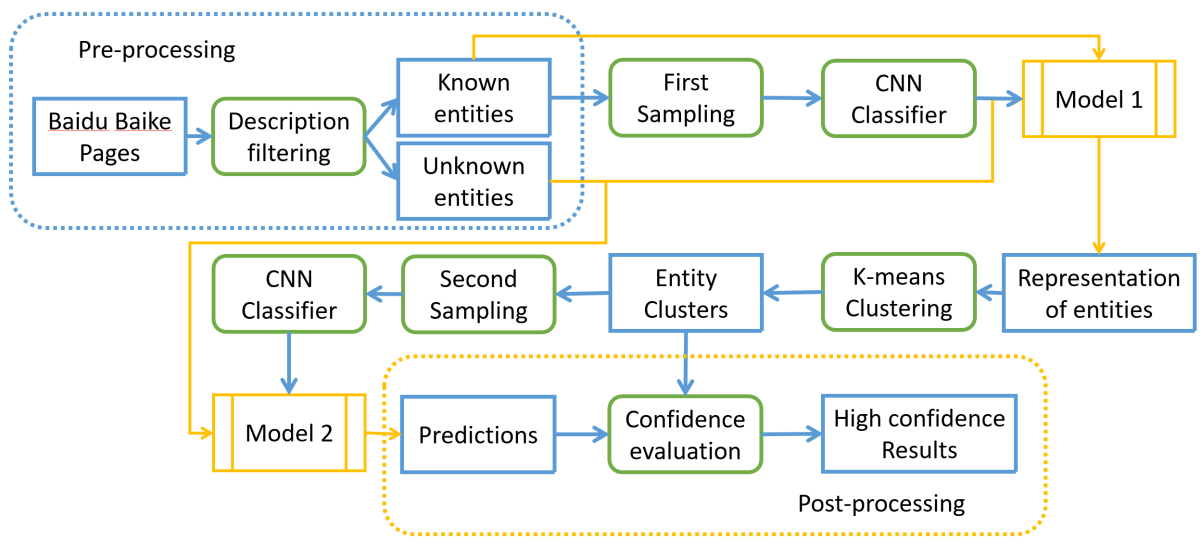

Figure 1: The workflow of the identification system.

Pre-processing is introduced to select entities that contain descriptions. Kazama and Torisawa (2007) directly regard the first sentence of an entity page as its description. However, we estimate from a sampled data of Baidu Baike and find that only approximately 84% of entity pages hold such assumption. To increase this percentage, we only regard the first sentence of an entity page as its description when it meets at least one of two requirements:

- The first sentence begins with the entity name (the title of entity page); or
- The head of the first sentence in its dependency tree is a verb phrase;

The unsatisfied entity pages are directly filtered. After filtering, the precision of the descriptions is increased to 97%. Nearly 9.7 million and 2.1 million entities are reserved from known and unknown entity sets, respectively, which constitute our training and prediction set. The training set has two problems that need to be addressed. First, since the precision of current entity-hypernym relationships is approximately 98%, the other 2% should be regarded as the noise, which would lead to performance decline for a non-specially designed model (Zhu and Wu, 2004). Second, since different concepts (classes) correspond to different numbers of entities (see Table 1), the training set is imbalanced, which would cause poor performance in minority classes. Instances in minority classes would also suffer from higher noise.

We adopt a standard convolutional neural network (CNN) classifier (Kim, 2014) for text classification. To solve the noise and imbalance problem, we first train the CNN model using small-sized balanced data. Based on the trained model, we obtain the vector representation of all training and predict entities. Then, we use a clustering-based re-sampling method to select the highly confident entities from the total training data, to obtain another large-sized training set. In this set, the noise and imbalance problems are alleviated. We re-train the CNN classifier accordingly and obtain the final classification model, which is used to predict the hypernyms of unknown entities.

To increase the precision of the prediction, we introduce a post-processing module to select the highly confident predicted entities. The details of the CNN classifier, the clustering and re-sampling module, and the post-processing module are described below.

## 2.3 CNN classifier

As shown in Figure 2, the CNN classifier has two input channels: the character-level name information and the word-level descriptive sentences, which relies on a word segmenter. We add special <start> and <end> symbols to both sides of the entity name, to help the CNN model capture the prefix and suffix features. The name information is removed from the description, to avoid feature redundancy. If the name is surrounded by the title mark (≪≫), a special Chinese punctuation to indicate the titles of books, films, and more, it would be moved along with the title to the name channel.

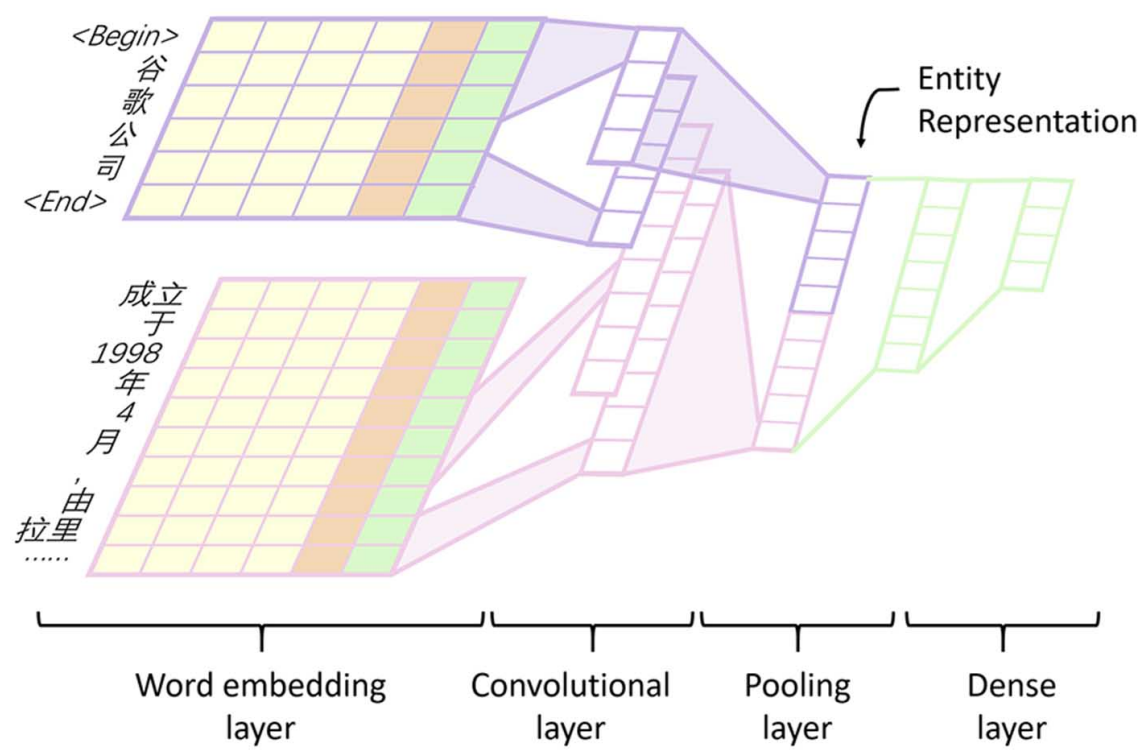

Figure 2: The architecture of the CNN classifier. The input of the name channel is "Google Inc.", while that of the description channel is "established in April, 1998 by Larry...".

The embedding of words has dimensionality of 200, which are pre-trained from Baidu Baike articles using the skip-gram model (Mikolov et al., 2013). The character-level and word-level embedding tables are shared, because the single character word is common in Chinese. Since the skip-gram model has a threshold to ignore less frequent words, these words do not have pre-trained embedding and would be assigned to random embedding. A common way to enhance the word embedding is to concatenate their corresponding part-of-speech (POS) and syntax embedding (Wang and Chang, 2016; Chen and Manning, 2014). These features are reported to be beneficial to the hypernym identification. We therefore concatenate them to the word embeddings (shown in Figure 2 with different colors). The POS tagger, dependency parser, and the earlier segmenter are supported by the Baidu NLP Cloud service. Followed by a convolutional layer, a max-pooling layer, and two dense layers, we obtain the probability distribution of the input entity over the 48 classes.

### 2.4 Clustering and re-sampling

We first sample a relatively small number of entities randomly from each class and train a CNN model based on this small balanced set. The output of the pooling layer is a concatenated vector, and can be regarded as the representation of the entity. We expect that the entities in the same class would be close in the representation space. In other words, when we run a clustering algorithm over these entities, the cluster partition should be similar with the inherent classifications. If the percentage of a class in one cluster is tiny (e.g., less than $p\%$), we would suspect the entities in this class as noise and remove them. This would cause false filtering, but it would be better to remove several correctly labeled entities, rather than keep the falsely labeled ones. In this way, part of noise can be discarded.

The imbalanced problem is still alleviated with the undersampling strategy. Balanced training data is fair for minority classes, while imbalanced data is better for the majority. To compromise, the size of sampled data in each class is still correlated with that of entire training data, but the proportion is smoothed by the logarithmic function. We first set a reference sample size $N$. For one class $c$, we denote the number of entities in $c$ by $P_c$ before noise filtering, and $Q_c$ after filtering, where $P_c \geq Q_c$. There are three conditions (All the entities below refer to those belonging to class $c$):

1. If $P_c > N$ and $Q_c \geq N$, we would completely sample $\hat{N}_c = N \times (1 + \log_{10}(\frac{Q_c}{N}))$ entities from $c$;
2. If $P_c > N$ but $Q_c < N$, we rank the clusters in descending order by the number of filtered entities in $c$, and recall these filtered entities in order to make $Q_c$ exactly larger or equal to $N$. It then becomes the first condition;
3. If $P_c < N$, we directly take all the entities (filtered and remained) to the sampled data.

The final problem is the sampling method. It relies on the clustering results we obtained. Instances in the large-sized class would be partitioned into many small clusters. If we under-sample in the class-level, we have the risk of not sampling entities from small-sized clusters, and may lead to the wrong classification for the predict entities in the same cluster. We therefore sampled from the cluster level in a proportional way, which can better guarantee that the sampled subset can cover all clusters. An example is shown in Figure 3, which compares the results of sampling from the class or cluster level.

### 2.5 Post-processing

To add predicted data into current taxonomy, we have to guarantee the precision benchmark of 95%, which is hard to achieve. We must select the highly confident predicted entities from the entire prediction set. Similar to the methods to detect noise data, this process also relies on the clustering results.

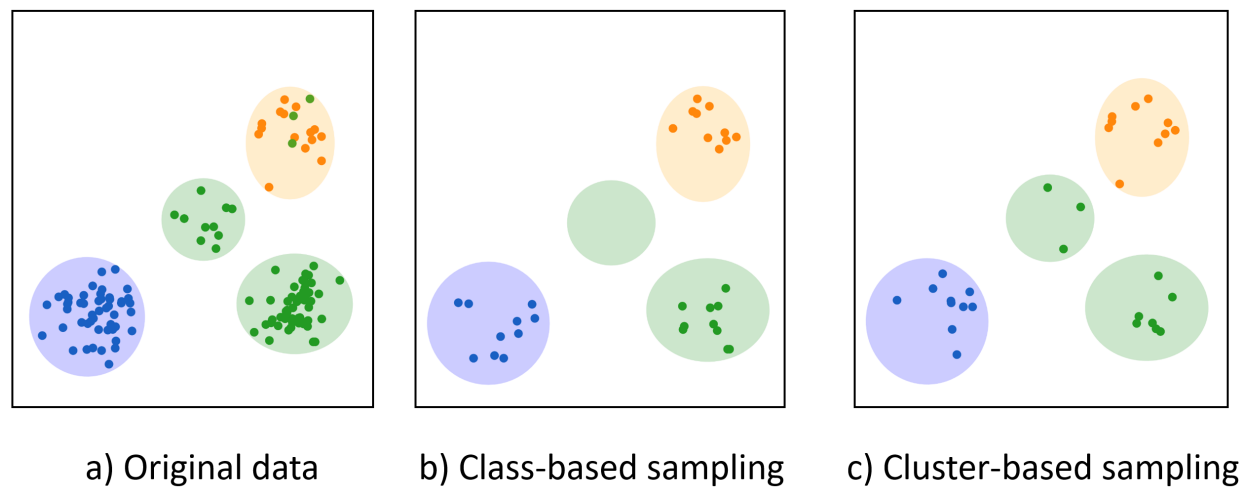

Figure 3: Resampling from the clusters. The ellipses represent the clusters and the color refers to the class. The noise is first removed from the original data. The class-level sampling fails to sample data from the center cluster, while the cluster-level sampling avoids this situation. The sampling process is simulated via the random.sample() method in Python.

In most cases, one cluster has both known and unknown entities. As shown in Figure 4, we can easily obtain the class distribution of known entities. The motivation of the confidence evaluation is intuitive: If most of the known entities are labeled as ★, and a test entity in the same cluster is also labeled as ★ by the classifier, then we have high confidence that this entity is correctly labeled. We grade the predict confidence of an unknown entity as four levels, according to the percentage of this class in the known entities: L1 (>99%), L2 (99%-50%), L3 (50%-5%), and L4 (<5%). If there are only test entities in a cluster, we simply label the confidence of a test entity as L5 if there are more than 50% entities labeled as the same class with it, or L6 otherwise.

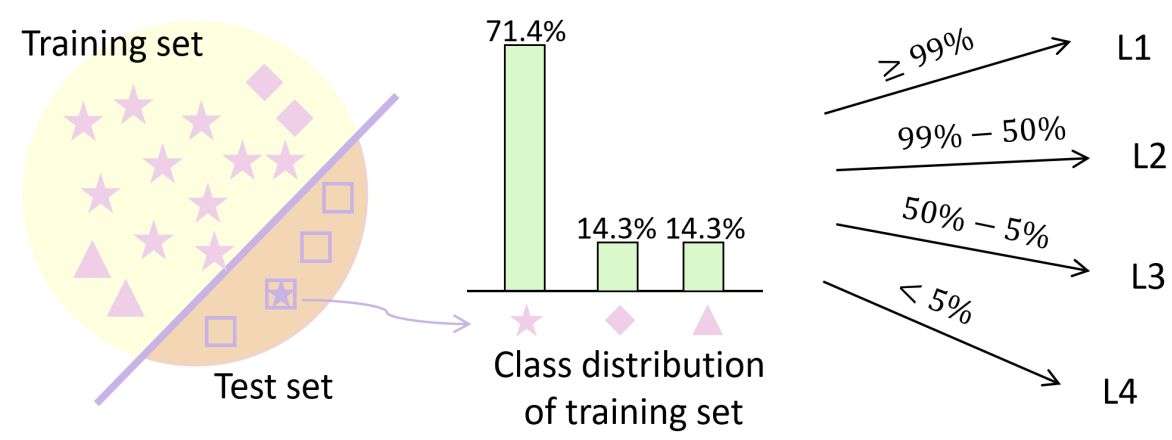

Figure 4: The confidence evaluation according to the class distribution of known entities. The unknown entities are represent with □, while other symbols represent the classes of known entities. Symbol in the □ is the prediction result of the unknown entity.

## 3 Experiment

### 3.1 The first classifier

To train the first classifier with the small balanced dataset, we randomly sampled 1,000 entities from each class, and split the training/validation set with ratio of 0.7/0.3 (the numbers of entities are 33,600/14,400). Since parameter tuning was the only purpose, there was no separate test set. The structure and parameter were set as follows, according to the performance of the validation set:

The convolutional layer used filters with the window size as 1, 2, 3, 4, and each size contained 60 feature maps for every channel. The dimension of the pooling layer, therefore, was 480, and that of the hidden layer was 200. Two dropout layers with rates of 0.5, as well as an $L_2$-norm constraint for dense-layer parameters were utilized for regularization. Rectified linear units (ReLUs) were adopted as the activation function of the convolutional layer and the first dense layer. The softmax cross entropy was used as the loss function. The optimizer adopted the Adam algorithm Kingma and Ba (2014) with a learning rate of 0.001 and a mini-batch size of 500. The performance of the trained classifier on each class of the validation set is shown in Table 1. All macro $P$, $R$, and $F_1$ measures are 0.88.

We evaluated the influence of POS and syntax features on the final performance. The introduction of POS features increased the macro-$F_1$ from 87.95% to 88.36%. However, the continued concatenation of dependency tagging features decreased the measure to 88.06%. Similar results were reported by Sang and Hofmann (2009). It may be caused by the relatively low precision of the Chinese dependency parser. Therefore, we only concatenated the POS tagging features to the word embedding.

We also tried to change CNN to long short-term memory networks (LSTMs), to combine the word and character sequences as representation vectors. We found that, neither replacing CNN with LSTM on the name channel nor description channel can cause a large slump of macro-$F_1$ measure by more than 10%. A possible reason is that the LSTMs without the attention mechanism focus more on the last part of the sentence, where the information for hypernym identification may not exist.

## 3.2 Clustering

With the help of the classifier, we obtained the vector representation of all entities, with the dimensionality of 480. We run the K-means algorithm to partition these entities, and $K$ is arbitrarily set as 10,000. The statistical quantities of these clusters are shown in Figure 5. Figure 5 a) shows that the number of clusters which possess more than 1,000 entities is roughly equal to that of clusters with less than 1,000 entities. Figure 5 b) and c) indicate that approximately one half of the clusters are almost pure, in which more than 99% of the entities belong to the same class.

We further analyzed the clusters of different classes. For each class, we select the clusters where this class has the most number of entities, and draw the boxplot of these numbers. The results are shown in Figure 6. The large-sized classes are more likely to be partitioned into large clusters. This indicates that despite the large numbers, the descriptions of these entities in the same class are not arbitrary. They instead have high similarity, and therefore the classification-based method is reasonable.

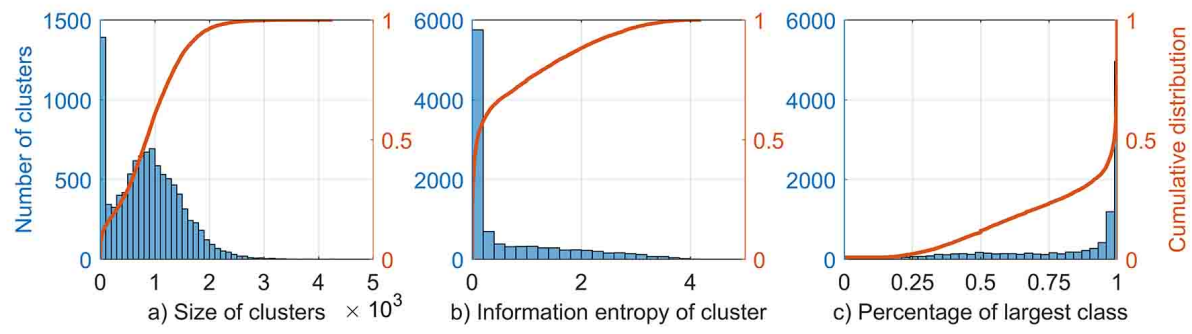

Figure 5: a) and b) show the distribution of size and information entropy of clusters, and c) shows the distribution of the percentage of the largest class in one cluster. All of the left y-axes represent the number of clusters, and right y-axes represent the corresponding cumulative distribution.

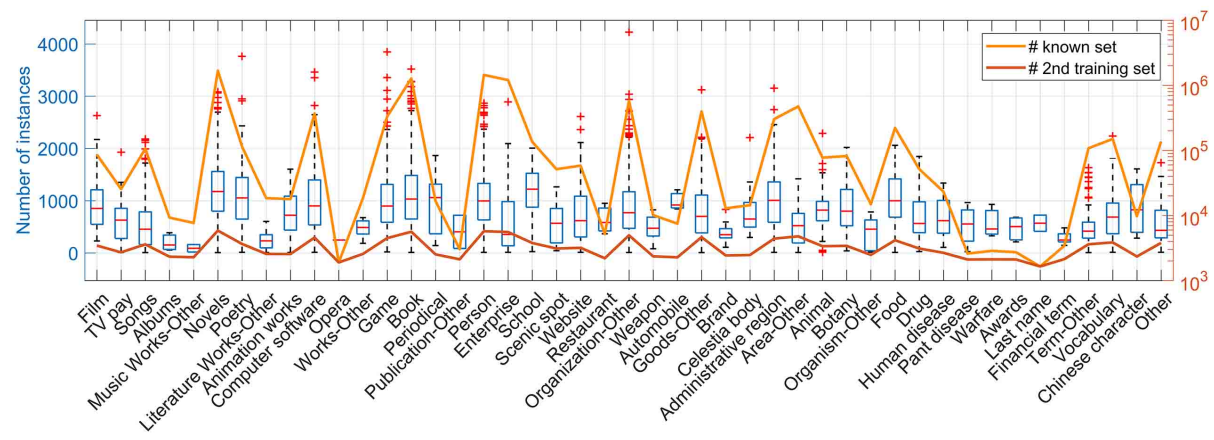

Figure 6: The boxplot described above. We also draw the number of entities in the known set, as well as the second training set for comparison. These refer to the right log-scaled axis.

## 3.3 The second classifier

We re-sampled training entities from these clusters using the strategy described in Section 2.4. Compared with the previous training set, the size of this re-sampled set enlarged from 33,600 to 154,479 (see Table 1). It can better represent the entire training set. Based on this set, we re-trained the second classifier from scratch, and then applied it to predict the hypernym of unknown entities. The post-processing module then partitioned these entities into six confidence levels. The precision of the predicted results was evaluated manually, by sampling a small subset. For simplicity, we refer to a confidence level of one class as a *group*. We randomly selected at least 40 entities from each group and labeled the predicted hypernyms as right or wrong. More than 10,000 entities are evaluated in total. In particular, entities being classified as "Other" would be linked to the higher-level concepts. For example, if a film is classified as VIDEOWORKS-OTHER, we would regard it as a VIDEO WORK and therefore labeled this prediction as correct.

The precision of each class $c$ in the entire predict set was estimated by two ways:

1. Estimation at the class level: we sampled $m$ entities from class $c$ for manual evaluation in all, and $r$ of them were right identified. The precision was calculated by $p_1 = r/m$;

2. Estimation at the group level: we denote by $n$ the size of class $c$. Its $i$-th confidence level $L_i$ has $n_i$ entities with precision of $p_i$, which was estimated by the first way. Then $p_2 = (\sum_i p_i \cdot n_i)/n$

Since the evaluation subset was not uniformly sampled, we regarded $P_2$ as the primary estimation of the class-level precision. The size of some groups was less than 100, which would make small contributions to the expansion of our taxonomy. Therefore, we did not evaluate them.

We tried to correct the wrongly labeled entities while evaluating but found it difficult and time-consuming in practice. Therefore, the recall of the results could not be evaluated. Fortunately, we were more concerned about the precision, and the absence of recall was acceptable.

# 4 Results and discussion

## 4.1 Results

Table 1 shows the details of the precision of each group. Generally speaking, the precision declines from level 1 to level 4. Level 1 has a perfect performance, because they are selected from the almost pure clusters. The number of entities belonging to this level, as a result, is not very much, except several classes like ENTERPRISE, PERSON and POETRY. Instances of other concrete classes concentrated mainly on level 2 and level 3. There are still 49.7% of the classes in level 2 with a precision of more than 94%, and this percentage decreases to 8.7% in level 3. The special "Other" classes, which contain the remaining concepts not listed in the taxonomy, center on the fourth level with a relatively lower precision. It is because these "Other" classes contain more messy concepts, and are unlikely to be clustered into the relatively pure groups.

Instances in the fifth and sixth levels come from the clusters consisting of only predict entities, and there are two situations. If it is an accidental consequence caused by the K-means algorithm, the precision of this cluster would be affected by its nearest clusters. If it is caused because no similar entities existing in the known set, the precision would be low.

Finally, we choose the groups in which the precision is more than 94%, and add these entities to our existed taxonomy. 1.1 million out of 2.1 million are selected in this way, and the percentage is approximately 55%. According to the second precision estimation method, our selected subset has the precision of 99.36%.

## 4.2 discussion

The first sentence of an instance page from the web encyclopedia is an ideal source for the entity categorization task. The page editor would write the first sentence more clearly. In many cases, it is a standard definition sentence. On the other hand, when the editors try to write a new article, they may refer to the existing pages of the same kind of instances, leading to the similar form of sentences among these instances.

Since the precision of the knowledge base is more important, we designed the post-processing module to divide the predicted results into different confidence levels, and evaluate each level separately. In essence, it is a strategy to assign confidence to predicted instances according to the support number of similar training instances. The largest contribution to the high precision comes from the ENTERPRISE type, in which the names of entities have obvious suffix features (e.g., Inc.) and therefore a simple classifier can obtain satisfactory results. However, even we do not consider these entities, the precision of prediction set before and after the post-processing is 80.72% and 98.76%, respectively, which can still show the effectiveness of confidence-based filtering to the high precision.

The prerequisite of the classification strategy is the predefined taxonomy, which makes the categorization process more controllable compared with the hypernym extraction methods, but also restricts the granularity of the hypernym concept. How fine-grained should an identified category be depends on the subsequent tasks. Consider *Phenobarbital* as an example. For a medical ontology which is used for clinical support, it should be described as a hypnotics/anticonvulsants or in more detail. In other applications, like the intension detection of medical QA web, it is enough for the system to know it is a medicine, and therefore the coarse-grained categorization is acceptable.

## 4.3 Limitations

Despite the application background, it would be better to find a taxonomy that can compromise the granularity of the hypernym and the classification precision. In the clusters, we find that several clusters in the same class correspond to different sub-classes. The PERSON class, for example, is partitioned as ACTOR, SINGER, WRITER, and more, which shows the potentiality of the classifier for finer-grained classification.

Another limitation is the selection of descriptive sentences. Instead of regarding the first sentence of the Baidu Baike pages as the description, it is better to train a discriminator to select high-quality descriptions from web text. In this way, the entity set would not be restrained to those involved in Baidu Baike.

We utilized a simple CNN text classifier and flattened the hierarchical structure of the taxonomy during classification. More elaborate operations should further improve the performance.

# 5 Related works

## 5.1 Hypernym extraction

Hypernym extraction is utilized to find the hypernym of a certain entity from the sentences where the hypernym and the entity occur together. For example, from the sentence "... such authors as Herrick and Shakespeare." we can derive that "Herrick" and "Shakespeare" are authors. Such methods are mainly based on the lexical or syntactic patterns. They were designed manually at the beginning by Hearst and Hearst (1992), and then were mined automatically using machine learning techniques (Snow, Jurafsky, and Ng, 2004; Ritter, Soderland, and Etzioni, 2009; Kozareva and Hovy, 2010). These methods require the co-occurrence of hypernym-hyponym pairs in one sentence, and often suffer from relatively lower precision or recall (Seitner et al., 2016), because of the arbitrariness of the natural language.

| Taxonomy | | | validation set | | | predict set | | | level 1 | | level 2 | | level 3 | | level 4 | | level 5 | | level 6 | |
|---|---|---|---|---|---|---|---|---|---|---|---|---|---|---|---|---|---|---|---|---|
| | | | P | R | F | #num | $P_1$ | $P_2$ | #num | P | #num | P | #num | P | #num | P | #num | P | #num | P |
| Works | Video works | 1. Film | 0.81 | 0.95 | 0.88 | 8568 | 0.72 | 0.79 | 571 | 1 | 5764 | 0.94 | 883 | 0.32 | 1347 | 0.4 | | | 3 | |
| Works | Video works | 2. TV play | 0.89 | 0.9 | 0.9 | 3425 | 0.89 | 0.87 | | | 2086 | 0.94 | 835 | 0.9 | 504 | 0.5 | | | | |
| Works | Video works | 3. TV program | 0.88 | 0.99 | 0.93 | 5404 | 0.61 | 0.73 | | | 363 | 0.52 | 4053 | 0.77 | 985 | 0.62 | | | 3 | |
| Works | Video works | 4. Other | 0.59 | 0.54 | 0.56 | 6831 | 0.77 | 0.77 | | | | | | | 6823 | 0.77 | | | 8 | |
| Works | Music works | 5. songs | 0.91 | 0.93 | 0.92 | 10271 | 0.86 | 0.97 | 3072 | 1 | 6378 | 1 | 603 | 0.65 | 218 | 0.33 | | | | |
| Works | Music works | 6. albums | 0.86 | 0.94 | 0.9 | 1575 | 0.73 | 0.80 | | | 940 | 1 | 471 | 0.52 | 163 | 0.45 | | | 1 | |
| Works | Music works | 7. Other | 0.79 | 0.72 | 0.76 | 27436 | 0.62 | 0.56 | | | | | | | 27423 | 0.56 | 5 | | 8 | |
| Works | Literature works | 8. novels | 1 | 1 | 1 | 8806 | 0.82 | 0.93 | 2302 | 1 | 5807 | 0.97 | 594 | 0.48 | 98 | | | | 5 | |
| Works | Literature works | 9. poetry | 0.91 | 0.97 | 0.94 | 8660 | 0.88 | 0.94 | 3531 | 1 | 3693 | 0.94 | 913 | 0.71 | 519 | 1 | | | 4 | |
| Works | Literature works | 10. Other | 0.76 | 0.68 | 0.72 | 30141 | 0.61 | 0.60 | | | | | 21455 | 0.65 | 8686 | 0.49 | | | | |
| Works | 11. Animation works | | 0.88 | 0.85 | 0.87 | 6266 | 0.73 | 0.70 | 5 | | 2741 | 1 | 1401 | 0.61 | 2116 | 0.38 | | | 3 | |
| Works | 12. Computer software | | 0.9 | 0.95 | 0.93 | 12670 | 0.74 | 0.63 | 203 | 1 | 6274 | 0.87 | 4892 | 0.35 | 1294 | 0.5 | | | 7 | |
| Works | 13. Opera | | 0.9 | 0.79 | 0.84 | 6362 | 0.68 | 0.55 | | | 1418 | 1 | 1115 | 0.52 | 3825 | 0.39 | | | 4 | |
| Works | 14. Other | | 0.86 | 0.68 | 0.76 | 15957 | 0.39 | 0.33 | | | | | | | 14760 | 0.3 | 898 | 0.77 | 299 | 0.4 |
| 15. Game | | | 0.94 | 0.94 | 0.94 | 12562 | 0.78 | 0.72 | 1200 | 1 | 5214 | 1 | 4991 | 0.32 | 1156 | 0.9 | | | 1 | |
| Public-ations | 16. Book | | 0.78 | 0.89 | 0.83 | 34347 | 0.62 | 0.77 | 6718 | 1 | 21072 | 0.77 | 4956 | 0.45 | 881 | 0.92 | 119 | 1 | 601 | 0.6 |
| Public-ations | 17. Periodical | | 0.94 | 0.98 | 0.96 | 2264 | 0.81 | 0.89 | 6 | | 1609 | 0.94 | 230 | 0.65 | 418 | 0.83 | | | 1 | |
| Public-ations | 18. Other | | 0.93 | 0.81 | 0.86 | 2827 | 0.71 | 0.71 | | | | | | | 2826 | 0.71 | | | 1 | |
| 19. Person | | | 0.86 | 0.95 | 0.9 | 289215 | 0.76 | 0.94 | 215230 | 1 | 63179 | 0.78 | 8751 | 0.55 | 1964 | 0.65 | 11 | | 80 | |
| Organizations | 20. Enterprise | | 0.98 | 0.93 | 0.96 | 678962 | 0.68 | 0.99 | 624168 | 1 | 6043 | 0.94 | 2268 | 0.65 | 9993 | 0.94 | 35830 | 0.88 | 660 | 0.84 |
| Organizations | 21. School | | 0.95 | 0.98 | 0.97 | 6397 | 0.8 | 0.90 | 2275 | 1 | 2116 | 1 | 1221 | 0.55 | 769 | 0.88 | | | 16 | |
| Organizations | 22. Scenic spot | | 0.83 | 0.95 | 0.89 | 15836 | 0.96 | 0.91 | 192 | 1 | 2951 | 1 | 10418 | 0.87 | 2271 | 1 | | | 4 | |
| Organizations | 23. Website | | 0.92 | 0.94 | 0.93 | 22456 | 0.62 | 0.66 | 16 | | 12792 | 0.84 | 4743 | 0.45 | 4898 | 0.39 | | | 7 | |
| Organizations | 24. Restaurant | | 0.91 | 0.94 | 0.93 | 13521 | 0.97 | 0.96 | | | 3047 | 1 | 7409 | 0.97 | 3065 | 0.89 | | | | |
| Organizations | 25. Other | | 0.83 | 0.69 | 0.75 | 79195 | 0.69 | 0.96 | | | | | | | 77129 | 0.97 | 1855 | 0.45 | 211 | 0.36 |
| Goods | 26. Weapon | | 0.9 | 0.96 | 0.93 | 8435 | 0.85 | 0.76 | 56 | | 2334 | 0.9 | 2920 | 0.61 | 3113 | 0.82 | | | 12 | |
| Goods | 27. Automobile | | 0.92 | 0.94 | 0.93 | 5949 | 0.79 | 0.69 | | | 1294 | 0.94 | 2772 | 0.71 | 1878 | 0.5 | | | 5 | |
| Goods | 28. Other | | 0.82 | 0.86 | 0.84 | 94750 | 0.85 | 0.85 | | | | | | | 94700 | 0.85 | | | 50 | |
| 29. Brand | | | 0.84 | 0.85 | 0.84 | 25332 | 0.72 | 0.70 | | | 6733 | 0.97 | 9821 | 0.68 | 8768 | 0.52 | | | 10 | |
| 30. celestial body | | | 0.99 | 0.98 | 0.98 | 6799 | 0.98 | 0.99 | 3322 | 1 | 1048 | 0.94 | 775 | 1 | 1654 | 1 | | | | |
| Area | 31. Administrative region | | 0.97 | 0.92 | 0.95 | 13447 | 0.69 | 0.96 | 8975 | 1 | 3131 | 1 | 632 | 0.71 | 538 | 0.5 | 138 | 1 | 33 | |
| Area | 32. Other | | 1 | 1 | 1 | 94760 | 0.94 | 0.92 | | | | | | | 94653 | 0.92 | 78 | | 29 | |
| Orga-nism | 33. Animal | | 0.86 | 0.96 | 0.91 | 18658 | 0.95 | 0.89 | 341 | 1 | 10685 | 0.94 | 5146 | 0.97 | 2479 | 0.5 | | | 7 | |
| Orga-nism | 34. Botany | | 0.88 | 0.89 | 0.88 | 22962 | 0.82 | 0.75 | 204 | 1 | 10939 | 0.9 | 7987 | 0.65 | 3818 | 0.5 | | | 14 | |
| Orga-nism | 35. Other | | 0.91 | 0.8 | 0.85 | 6723 | 0.94 | 0.94 | | | | | | | 6721 | 0.94 | | | 2 | |
| 36. Food | | | 0.91 | 0.89 | 0.9 | 12063 | 0.83 | 0.80 | 1515 | 1 | 6908 | 0.9 | 1975 | 0.65 | 1662 | 0.4 | | | 3 | |
| 37. Drug | | | 0.92 | 0.87 | 0.89 | 23576 | 0.7 | 0.60 | 585 | 1 | 11459 | 0.84 | 8150 | 0.29 | 3377 | 0.5 | | | 5 | |
| 38. Chemical substance | | | 0.8 | 0.9 | 0.84 | 21296 | 0.97 | 0.97 | | | | | 5983 | 0.97 | 15306 | 0.97 | | | 7 | |
| 39. Human disease | | | 0.93 | 0.9 | 0.92 | 10844 | 0.61 | 0.51 | 20 | | 4544 | 0.74 | 3984 | 0.39 | 2292 | 0.26 | | | 4 | |
| 40. Plant disease | | | 0.97 | 0.99 | 0.98 | 659 | 0.81 | 0.79 | 1 | | 519 | 1 | 40 | | 99 | | | | | |
| 41. Warfare | | | 0.97 | 0.96 | 0.96 | 2198 | 0.47 | 0.49 | | | 930 | 0.71 | 430 | 0.26 | 838 | 0.36 | | | | |
| 42. Awards | | | 0.98 | 0.98 | 0.98 | 2654 | 0.73 | 0.82 | | | 1728 | 0.97 | 714 | 0.55 | 208 | 0.48 | | | 4 | |
| 43. Last name | | | 0.96 | 0.92 | 0.94 | 2281 | 0.83 | 0.78 | | | 585 | 0.87 | 678 | 0.81 | 1018 | 0.71 | | | | |
| Term | 44. Financial term | | 0.88 | 0.96 | 0.92 | 8157 | 0.99 | 0.96 | | | 857 | 1 | 5379 | 0.97 | 1909 | 0.94 | | | 12 | |
| Term | 45. Other | | 0.74 | 0.64 | 0.69 | 328667 | 0.59 | 0.65 | | | | | 121277 | 0.91 | 187041 | 0.49 | 20092 | 0.64 | 257 | 0.38 |
| 46. Vocabulary | | | 0.83 | 0.86 | 0.85 | 99821 | 0.9 | 0.92 | 7813 | 1 | 59575 | 0.99 | 23002 | 0.81 | 9396 | 0.71 | | | 35 | |
| 47. Chinese character | | | 0.94 | 0.95 | 0.94 | 4386 | 1 | 0.99 | | | 4336 | 1 | 5 | | 45 | | | | | |
| 48. Other | | | 0.62 | 0.58 | 0.60 | | | | | | | | | | | | | | | |

Table 1: The statistical quantities and evaluation results of each class. From left to right, this table shows 1) the pre-defined concept taxonomy; 2) the P, R, and F measure of the first classifier on the validation set; 3) the number of entities predicted by the second classifier, as well the two estimated precision; and 4) the numbers and precision information of each group. The red shadow is a visualization of some important evaluation measures, while the green shadow indicates the percentage of the predicted entities in each confidence level.

### 5.2 Named entity recognition and classification

NERC was first introduced at the Sixth Message Understanding Conference (MUC-6) Grishman and Sundheim (1996). Early tasks focused most on three kinds of entity types: person, location, and organization. The classes were further fine-grained in subsequent works. For example, Fleischman and Hovy (2002) classified the person into sub-classes like politics and artist. In the field of biomedical informatics, researchers would focus more on the identification of specific terms, like proteins and DNAs (Leaman, Gonzalez, and Others, 2008). Nadeau (2007) presented a hierarchical structure for named entities, and the number of entity types is approximately 200. Ling and Weld (2012) defined 112 fine-grained tags and labeled the training data with the help of anchor links from Wikipedia text, then utilizing the perceptron as the classifier to determine the type of entities. The dominant supervised methods for NERC extract features from tokens themselves and their contexts, and no exterior knowledge source is involved.

## 6 Conclusion

This work offers a classification-based method for entity categorization. Based on the name and description information of an entity, we can classify it to one of the types via a CNN classifier. A clustering module is designed for noise filtering, training set sampling, and confidence evaluation for predicted results. We applied this method to 2.1 million open-domain entities, and 1.1 million are successfully classified with a precision of 99.36%, demonstrating the efficiency of our method.

In the future, we hope to develop a discriminator to find the entity descriptions in broader knowledge sources, and to identify the type of an entity from multiple descriptions.